\documentclass[conference]{IEEEtran}
\ifCLASSINFOpdf
\else
\fi

\usepackage{algorithm}
\usepackage[noend]{algpseudocode}
\usepackage{amsmath}
\usepackage{graphicx}

\usepackage{booktabs}
\usepackage{multirow}
\usepackage{longtable}
\usepackage{dblfloatfix} 

\begin{document}
\title{Masking Neural Networks Using Reachability Graphs to Predict Process Events}

\author{\IEEEauthorblockN{Julian Theis and Houshang Darabi}
\IEEEauthorblockA{Mechanical and Industrial Engineering\\
University of Illinois at Chicago\\
Chicago, Illinois 60607\\
Email: \{jtheis3, hdarabi\}@uic.edu}}


\maketitle

\begin{abstract}
\textit{Decay Replay Mining} is a deep learning method that utilizes process model notations to predict the next event.
However, this method does not intertwine the neural network with the structure of the process model to its full extent.
This paper proposes an approach to further interlock the process model of \textit{Decay Replay Mining} with its neural network for next event prediction. The approach uses a masking layer which is initialized based on the reachability graph of the process model. 
Additionally, modifications to the neural network architecture are proposed to increase the predictive performance.
Experimental results demonstrate the value of the approach and underscore the importance of discovering precise and generalized process models. 
\end{abstract}


%
\IEEEpeerreviewmaketitle

\section{Introduction}\label{sec:introduction}
Process mining is a set of techniques that focuses on deriving insights into the behavior of various real-world systems. Execution runs of a system are recorded in event logs and are fed to algorithms to discover explainable process models \cite{van2012process, processmining}. Ideally, the resulting process models describe the behavior observed in the event log and infer knowledge about potential future execution runs.
An example system can be \textit{patient care} in a hospital where an execution run consists of the observation of the steps, i.e. events, that a patient undergoes. An execution run starts with the observation of the \textit{patient admission} and ends with the \textit{patient discharge} event. The possible set of events is usually finite and limited by the nature of the system. In the example case, this set can encompass events such as \textit{lab measurement}, \textit{surgery}, \textit{triage}, or \textit{ICU transfer}. Process mining algorithms then can be applied to discover a process model that provides insights on e.g. inefficiencies of the system. The outcomes can help the administration to manage resources efficiently or assist physicians in decision-making. Nowadays, process mining applications can be found across a wide spectrum of industries and are becoming increasingly popular in commercial solutions.

Apart from discovering process models from event logs, decision-makers are interested in predictive process analytics. This includes the prediction of future events during process execution to detect unexpected deviations and to allow for timely intervention. Over the past years, several different deep learning methods have been developed to predict the next event \cite{pasquadibisceglie2021multi, litrev_tax, taymouri2020predictive}. 
These deep learning algorithms demonstrate superior predictive performance but are commonly trained using subsequences of events without incorporating knowledge resulting from process mining efforts. Only a few approaches enrich process mining methods with deep learning-based predictive abilities \cite{pasquadibisceglie2021multi}. Hence, most deep learning-based next event prediction algorithms are decoupled from process discovery outcomes. This seems to be counterintuitive as the discovery of process models is used to gain knowledge about the underlying system \cite{polyvyanyy2021bootstrapping}. Therefore, using process models in the task of predicting the next event is supposed to be beneficial.

One of the few prediction techniques that utilize process mining models for the next event prediction task is \textit{Decay Replay Mining} \cite{dreamnap}. This method extends structural components of a process model with time decay functions. 
Sequences of events can be replayed to obtain vectors of process model states which are used to develop a neural network that predicts the next event. The value of \textit{Decay Replay Mining} has been demonstrated in applied research works \cite{Theis2019c, diabetesicu}. 
Although utilizing the underlying process model, the neural network is not interlocked with the process model to its full extent. 

The contribution of this paper is an extension of \textit{Decay Replay Mining} to increase the incorporation of process mining outcomes into the task of the next event prediction. 
Specifically, the neural network is interlocked with the process model by masking the probabilistic outputs using the reachability graph of the process model. 
Additionally, neural network architecture modifications are proposed.
The approach is experimentally tested on benchmark datasets that were used to develop the original \textit{Decay Replay Mining} method.

The paper is structured as follows. 
After the introduction in Section \ref{sec:introduction},  preliminaries are introduced in Section \ref{sec:preliminaries}. Afterwards, related work is provided in Section \ref{sec:related-work}. 
Section \ref{sec:methodolgy} describes the proposed methodology followed by the experimental evalation in Section \ref{sec:experiments}.
Section \ref{sec:conclusion} concludes this paper.

\section{Preliminaries}\label{sec:preliminaries}
This section describes the mathematical concepts and notations that are required throughout the paper.

\subsection{Petri Nets}
The subsequent definitions are based on \cite{processmining} and \cite{dreamnap}. 
A Petri net (PN) is a mathematical model that is commonly used to represent processes.
It is defined using three sets: A set of places $\mathcal{P}$, transitions $\mathcal{T}$, and arcs $\mathcal{F} \subseteq (\mathcal{P} \times \mathcal{T}) \cup (\mathcal{T} \times \mathcal{P})$. Places and transitions, i.e.  $\mathcal{P} \cup \mathcal{T}$, are nodes of a PN and can be unidirectionally connected using arcs. 
Transitions usually correspond to events and vice versa. A labeled PN describes that the transitions correspond to events. The formal definition is provided in Equation \ref{eq:pn}.
\begin{eqnarray}\label{eq:pn}
PN = \langle\mathcal{P}, \mathcal{T}, \mathcal{F}, \mathcal{A}, \pi\rangle
\end{eqnarray}
$\mathcal{A}$ denotes the set of all possible events of a system.  
The function $\pi$ maps an event to a transition and a transition to either an event of $\mathcal{A}$ or to a non-observable event $\perp$, i.e. $\pi: \mathcal{T} \rightarrow \mathcal{A} \cup \{\perp\}$. Hence, $\forall_{a \in \mathcal{A}} \exists ! _{t \in \mathcal{T}} \pi(t) = a$ applies.

A node $x$ is unidirectionally connected to another node $y$ iff $(x,y) \in \mathcal{F}$. This means that the \textit{incoming nodes} for any node $x \in \mathcal{P} \cup \mathcal{T}$ can be defined as $\bullet x = \{y | (y,x) \in \mathcal{F}\}$. Similarly, the \textit{outgoing nodes} can be defined as $x \bullet = \{y | (x,y) \in \mathcal{F}\}$.

Places can hold tokens. The function $\sigma(p)$ returns the non-negative integer number of tokens of a place $p \in \mathcal{P}$. The state of a PN is called a marking and is defined as a vector $M \in Z^{|\mathcal{P}|}$ where $\mathcal{Z}$ is the set of all non-negative integers. The value of $M_i$ corresponds to $\sigma(p_i)$ where $i$ is the $i$th place in $\mathcal{P}$ and $i = 1, ..., |\mathcal{P}|$. The set of all possible markings of a PN is denoted by $\mathcal{M}$.
In process mining, PNs have usually a dedicated source and sink place to indicate the start and end of a process. Any other node of the PN is located between those two nodes. Correspondingly, the initial state $M^{init}$ has a token in the dedicated source and the final state $M^{final}$ has a token in the dedicated sink place.

When executing a transition, the number of tokens of all incoming places of that transition is reduced by one whereas the number of outgoing places is increased by one. Formally, a transition can only be executed iff $\forall_{p \in \bullet t} \sigma(p) \geq 1$. If the condition holds, the transition is also called \textit{enabled}.
A special type of transition is the \textit{hidden transition} that maps to a non-observable event $\perp$. 

\subsection{Reachability Graphs}
A reachability graph of a PN is defined as $RG = \langle\mathcal{N}, \mathcal{K} \rangle$ where $\mathcal{N}$ is the set of nodes and $\mathcal{K} \subseteq (\mathcal{N} \times \mathcal{N})$ is a set of directed edges connecting nodes.
Each node in the $RG$ corresponds to exactly one marking in the PN. Hence, $\mathcal{N} = \mathcal{M}$.
An edge connecting a node $x$ to another node $y$ of the reachability graph represents the transition that needs to be executed to move from the corresponding marking of $x$ to the one of $y$. 
A function $\eta$ maps a node $N$ of an $RG$ to a marking $M$ of the PN. 
$\eta^{-1}$ is the inverse of the function and maps a marking $M$ of the PN to a node $N$ of the $RG$.
Similarly, the function $\kappa$ maps an edge of the RG, i.e. $K$, to the corresponding transition $T$ of the PN. 
The inverse function $\kappa^{-1}$ maps a transition $T$ to a set of edges of the $RG$.
Since an $RG$ of a PN is a one-to-one mapping, the conditions of Equations \ref{eq:cond_nodes} and \ref{eq:cond_trans} apply.
\begin{equation}\label{eq:cond_nodes}
    \begin{split}
        \forall_{N \in \mathcal{N} ~\exists! ~M \in \mathcal{M}}: M = \eta(N) \\
        \forall_{M \in \mathcal{M} ~\exists! ~N \in \mathcal{N}}: N = \eta^{-1}(M)
    \end{split}
\end{equation}
\begin{equation}\label{eq:cond_trans}
    \begin{split}
        \forall_{K \in \mathcal{K} ~\exists! ~T \in \mathcal{T}}: T = \kappa(K) \\
        \forall_{T \in \mathcal{T}}: |\eta^{-1}(T)| \geq 1
    \end{split}
\end{equation}

\section{Related Work}\label{sec:related-work}
The task of predicting the next event given a sequence of events has a long-standing history. Early algorithms relied mostly on explicit probabilistic methods such as Hidden Markov Models \cite{becker2014hmm},  Probabilistic Finite Automata \cite{breuker2016pfa}, and Probabilistic Process Models \cite{litrev_lakshmanan}. A more recent approach leverages Dynamic Bayesian Networks that result in competitive performance \cite{pauwels2020bayesian}. Such explicit probabilistic methods have the advantage to be explainable. Although the resulting predictive models do not necessarily use process model notations like PNs directly, they are usually relatable to process models. A disadvantage of such methods is that they commonly incorporate only a limited amount of event information and may be incapable of detecting long-term dependencies \cite{pauwels2020bayesian}. 

More recently, next event prediction algorithms are increasingly based on deep learning due 
the superior performance of neural networks. A body of literature has been proposed using multiple different deep learning architectures for predictive process analytics, such as 
recurrent neural networks \cite{litrev_evermann, litrev_tax}, 
convolutional neural networks \cite{al2018predicting, pasquadibisceglie2019using, di2019activity}, 
stacked autoencoders \cite{litrev_mehdiyev}, 
multi-view deep learning architectures \cite{pasquadibisceglie2021multi}, 
graph neural networks \cite{harl2020explainable}, and 
generative adversarial networks \cite{taymouri2020predictive}.
These deep learning models usually outperform probabilistic methods but are currently lacking interpretability issues since the processes are modeled implicitly using event log data.
Moreover, these approaches neglect knowledge of process models that were discovered due to earlier process mining efforts.

Only a few proposed approaches leverage deep learning and explicit process model notations to predict the next event. One of those approaches is \textit{Decay Replay Mining} \cite{dreamnap} which describes a fairly new method consisting of two steps. In the first step, the places of a given PN are enhanced with time decay functions. When replaying event sequences on the enhanced PN, one can obtain multiple vectors at any time $\tau$ of the sequence: A vector representing the current marking $M(\tau)$ of the PN, a vector $C(\tau)$ that represents the number of tokens that entered each place of the PN, and a vector that represents the value of the time decay function associated with each place, $F(\tau)$. The concatenation of those vectors is defined as a \textit{timed state sample} $S(\tau)$ and represents a time-based state representation of the PN in comparison to a time-independent state, such as a marking. The second step of \textit{Decay Replay Mining} is a feedforward neural network that takes a \textit{timed state sample} $S(\tau)$ as an input and outputs a probability vector over all events $\mathcal{A}$ of the process. 
The method has demonstrated statistically significant performance improvements and has found applications in different domains \cite{Theis2019c, diabetesicu}

\section{Methodology}\label{sec:methodolgy}
\begin{figure}[h]
  \begin{center}
    \includegraphics[width=250pt]{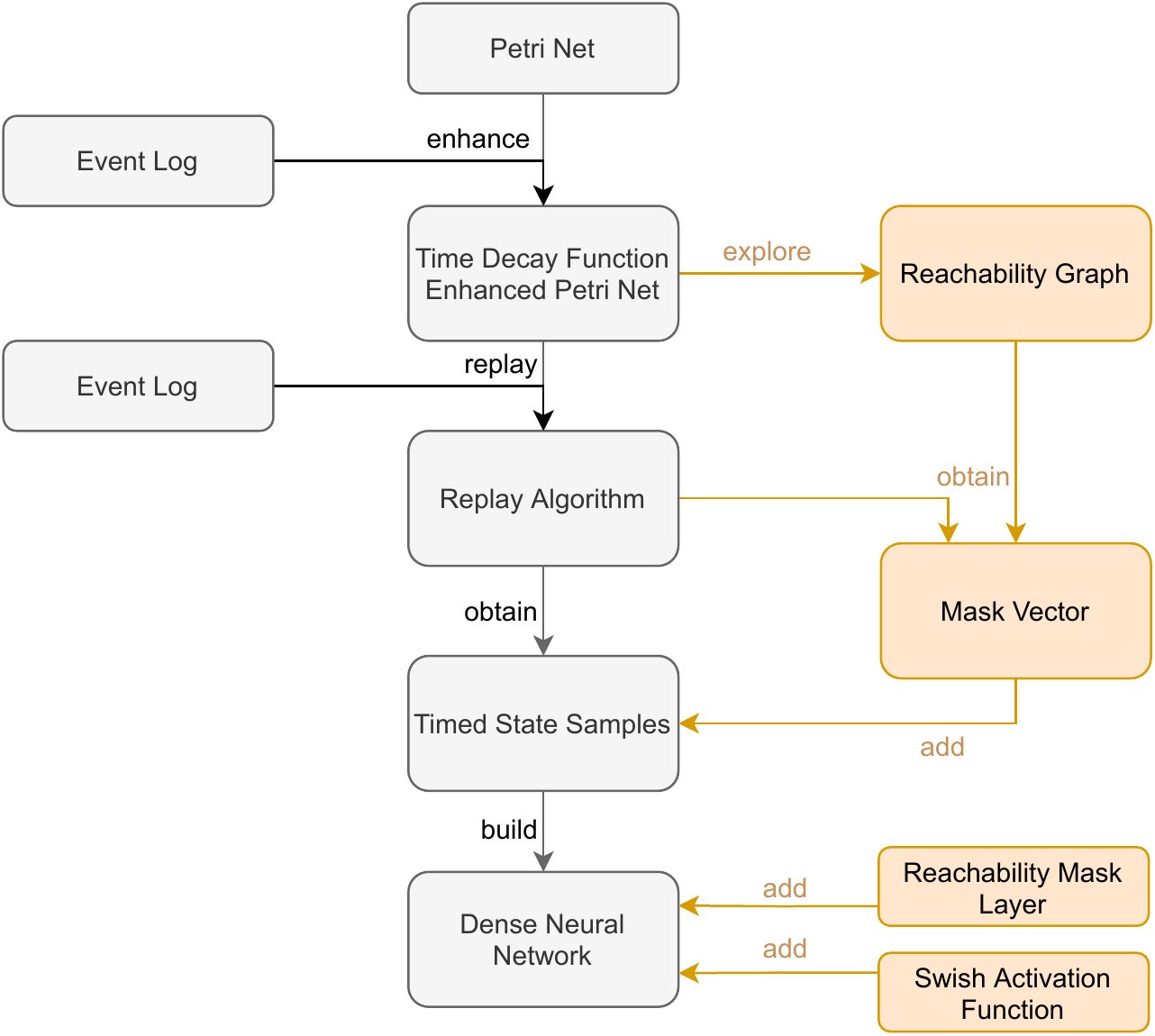}
  \end{center}
  \caption{Overview of the proposed adaptations of the \textit{Decay Replay Mining} approach. Blocks in gray represent the original components whereas orange ones highlighted the proposed adaptations.}
  \label{fig:overview}
\end{figure}

This section describes the extension of the \textit{Decay Replay Mining} method with a reachability masking functionality to increase the incorporation of the underlying PN into the process of predicting the next event. This extension assumes that the PN is a sound workflow net \cite{soundwfnet}. 
In total, three adaptations are proposed.
First, the PN enhancement is extended with a reachability graph exploration step. 
Second, masking vectors are obtained during replay and are added to the \textit{timed state samples}. 
Third, a reachability masking layer is added to the neural network and \textit{Rectified Linear Unit} activation functions \cite{relu} are replaced with \textit{Swish} activation functions \cite{ramachandran2017swish}.
Figure \ref{fig:overview} provides an overview of the proposed steps.

\subsection{Reachability Graph Exploration}
Every marking of the PN is a node in the reachability graph. All outgoing edges of such a node correspond to the possible next events that can occur, discovered by process mining algorithms. This characteristic is utilized to create subsets of $\mathcal{A}$ given a marking of the PN to exclude events that cannot occur according to the PN.
Since the reachability graph can be very large and edges might correspond to hidden transitions that map to non-observable events, a reduced reachability graph $RG_r$ is proposed.

$RG_r$ consists of a set of nodes $\mathcal{N}_{r} \subseteq \mathcal{N}$ and a set of edges $\mathcal{K}_{r}$. $RG_r$ is calculated such that $\mathcal{N}_{r}$ comprises $\eta^{-1}(M^{init})$ and nodes $N \in \mathcal{N}$ that fulfill Condition \ref{eq:cond_reducedrg}.
\begin{equation}\label{eq:cond_reducedrg}
    \forall_{K \in \bullet N}~ \pi\big( \kappa(K)\big) \neq \perp
\end{equation}
Given two nodes $N_1$ and $N_2$ such that $\{N_1, N_2\} \subset \mathcal{N}_r$ and $\eta^{-1}(M^{final}) \not\in \{N_1, N_2\}$, an edge $K_r \in \mathcal{K}_r$ connecting $N_1$ and $N_2$ exists iff 
a sequence of edges $\big(K_0, K_1, ..., K_n\big)$ where $n \geq 0$, $\{K_0, ..., K_n\} \subseteq \mathcal{N}$ exist such that
$N_1 \in \bullet K_0$, 
$N_2 \in K_n\bullet$, 
and $\forall_{i < n} \pi\big(\kappa(K_i)\big) = \perp$. This means that two nodes, i.e. markings of the PN, are connected if both markings can be reached via a visible transition or via a sequence of transitions where all transitions are hidden except the last one. 
Such an edge $K \in \mathcal{K}_r$ is then defined such that it maps to the  transition of the corresponding $PN$, i.e. $\kappa(K) = \kappa(K_n)$. Hence, all $K \in \mathcal{K}_r$ can be mapped to observable events that are contained in $\mathcal{A}$. This is the foundation for the construction of the later proposed reachability masks since a node of the $RG_r$ represents the marking of the PN that can be reached after executing a visible transition. Moreover, the outgoing edges express the set of executable visible transitions even if they are not enabled yet and require executing multiple hidden transitions first. 

The $RG_r$ is calculated after the time decay function enhancement of the $PN$. 
This $RG_r$ contains nodes for all markings that are reached immediately after executing a non-hidden transition. Non-hidden transitions correspond to events that are observed within an event log.
Moreover, outgoing edges of each node correspond to an $a \in \mathcal{A}$ and describes the next possible observable event according to the $PN$. 

\subsection{Mask Vectors}
Due to the properties of $RG_r$, one can easily detect the node $N \in \mathcal{N}_r$ that corresponds to the marking that is reached after replaying a sequence of events. 
The marking $M(\tau)$ that is contained in a timed state sample $S(\tau)$ is per definition the marking of the $PN$ that is reached after executing a transition that maps to an observable event.
Hence, $\eta^{-1}\big(M(\tau)\big) \in \mathcal{N}_r$. 
It is now obvious that the set of possible next events at a given marking $M(\tau)$ can be defined as $\mathcal{A}_{M(\tau)}$ such that 
$\forall_{K \in \eta^{-1}\big(M(\tau)\big)\bullet } \exists \pi\big(\kappa(K)\big) \in \mathcal{A}_{M(\tau)}$
and 
$|\eta^{-1}\big(M(\tau)\big)\bullet | = |\mathcal{A}_{M(\tau)}|$
according to the $PN$.

A masking vector $MSK$ is proposed which is a binary vector such that the length of $MSK$ equals to $|\mathcal{A}|$. 
A value of $1$ at the $i$th index of $MSK$ indicates the presence of the $i$th event of $\mathcal{A}$ in $\mathcal{A}_{M(\tau)}$. 
A value of $0$ indicates the non-presence, respectively. 
In this way, $MSK$ is a vector with a fixed length that describes the next possible events according to $PN$. 
If $MSK$ is a zero vector, then the final marking of the $PN$ is reached.

\subsection{Reachability Mask Layer}
This subsection describes how $MSK$ is used to mask the output of the \textit{Decay Replay Mining} neural network such that the resulting probabilities are set to $0$ for all $a \in \mathcal{A}-\mathcal{A}_{M(\tau)}$.
By following this step, the neural network is stronger interlocked with and incorporates knowledge of the process model.
Moreover, the prediction of the next event is reduced to a subset of the global event set of the system, i.e. to the set that is modeled by the PN which is supposed to reflect the system behavior. The neural network of \textit{Decay Replay Mining} originally consists of four fully-connected hidden layers where the $S(\tau)$ is the input to the first hidden layer and the output of the last hidden layer is a softmax layer providing a probability vector of the next event for each $a \in \mathcal{A}$.

The first proposed modification is the concatenation of the two vectors $S(\tau)$ and $MSK$ as an input to the neural network.
In this way, the hidden layers can learn and incorporate the potential next events in the early layers without the need to derive the information from $S(\tau)$ only.
This leads to presumably optimized latent representations in the hidden layers for the next event prediction.

The second proposed modification affects the output layer of the neural network. 
Rather than naively calculating the softmax probability vector over the global event set $\mathcal{A}$, the probability vector is calculated such that it forces to reduce the probabilities of all $a \in \mathcal{A}-\mathcal{A}_{M(\tau)}$ to zero. 
Therefore, the Hadamard product of the softmax output of the last neural network layer and $MSK$ is calculated. The resulting vector is then standardized to recover a probability vector which sums to $1$.
As a consequence, probabilities for the next event are only calculated over $\mathcal{A}_{M(\tau)}$ instead of the global event set $\mathcal{A}$.


\subsection{Swish Activation Function}
This paper proposes to replace the \textit{Rectified Linear Unit} activation functions of the neural network in \cite{dreamnap} with \textit{Swish} activation functions \cite{ramachandran2017swish}. The \textit{Swish} activation function has multiple potentially beneficial advantages compared to \textit{Rectified Linear Unit} function.
First, it is a smooth function and does not abruptly change the slope. Second, it is a non-monotonic function. 
As a consequence, the output landscape of the Swish activation function is smoother compared to the \textit{Rectified Linear Unit} function \cite{ramachandran2017swish}.
It can be assumed that \textit{Swish} provides an improved gradient signal to train the neural network of \textit{Decay Replay Mining} towards an optimum compared to \textit{Rectified Linear Unit} due to the sparse input features that are caused by the sparsity of the marking vector of the process model.


\section{Experimental Evaluation}\label{sec:experiments}
This subsection focuses on the evaluation of the proposed approach and is split into three parts.  
First, an overview over the experimental setup is provided in Section \ref{sec:eval:setup}.
Second, the impact of the proposed reachability-based masking on the predictive performance is experimentally evaluated in Section \ref{sec:eval:predictive}. 
Third, the role of precision is investigated in Section \ref{sec:eval:precision}

\subsection{Setup}\label{sec:eval:setup}
This subsection provides an overview over the leveraged datasets and the used metrics for comparison.

\subsubsection{Datasets}
The experimental evaluation is based on real-world benchmark datasets. Specifically, two \textit{Helpdesk} \cite{dataset_helpdesk, dataset_helpdeskC}, the \textit{Business Process Intelligence Challenge 2012} (BPIC'12) \cite{dataset_bpic12}, and the \textit{Business Process Intelligence Challenge 2013} (BPIC'13) \cite{dataset_bpic13} datasets are used. 

Both helpdesk event logs \cite{dataset_helpdesk, dataset_helpdeskC} originate from a ticketing management process of the helpdesk of an Italian software company.
The BPIC'12 \cite{dataset_bpic12} dataset originates from a Dutch financial institute and represents a loan application process. The event log is split into five subprocesses. Each one is used as a separate dataset for the experimental evaluation of the proposed approach. 
The first one contains all events that correspond to the \textit{work} subprocess with a lifecycle \textit{complete} status.
The second one contains all events that correspond to the \textit{work} subprocess independent of the lifecycle status.
The third one contains all events that correspond to the loan application \textit{offer}.
The fourth subprocess denotes the event log that encompasses all events that correspond to the subprocess of the \textit{application}. 
Finally, the last subprocess of BPIC'12 denotes all events contained in the event log.
The BPIC'13 dataset \cite{dataset_bpic13} originates from an incident and problem management system of Volvo IT. This dataset consists of two subprocesses.
One that denotes the subprocess of \textit{problem management} which consists of all corresponding events. The second subprocess corresponds to the \textit{incident management}. As before, each subprocess is used as a separate dataset. 

This leads to a total of 9 different datasets for experimental evaluation. Each one is randomly split into a training, validation, and testing subset using a respective 60/20/20 ratio.

\subsubsection{Metrics}
The evaluation of the proposed method is based on the unweighted and weighted multiclass Receiver Operating Characteristic Area Under the Curve (mAUROC) metrics. The unweighted mAUROC is calculated as the mean of the one-vs-rest AUROC scores for each $a \in \mathcal{A}$. The weighted mAUROC is the weighted mean of the one-vs-rest AUROCs of each $a \in \mathcal{A}$  using the support of events.

Each experimentally evaluated model is trained five times per dataset. This ultimately results in 45 different models for evaluation. 
Additionally, the originally proposed \textit{Decay Replay Mining} model is used for baseline comparisons. 
The differences of mAUROCs between the proposed model and the baseline model per dataset are used to calculate 95\% confidence intervals (CIs) of the performance differences.

\begin{figure}[b]
  \begin{center}
    \includegraphics[width=250pt]{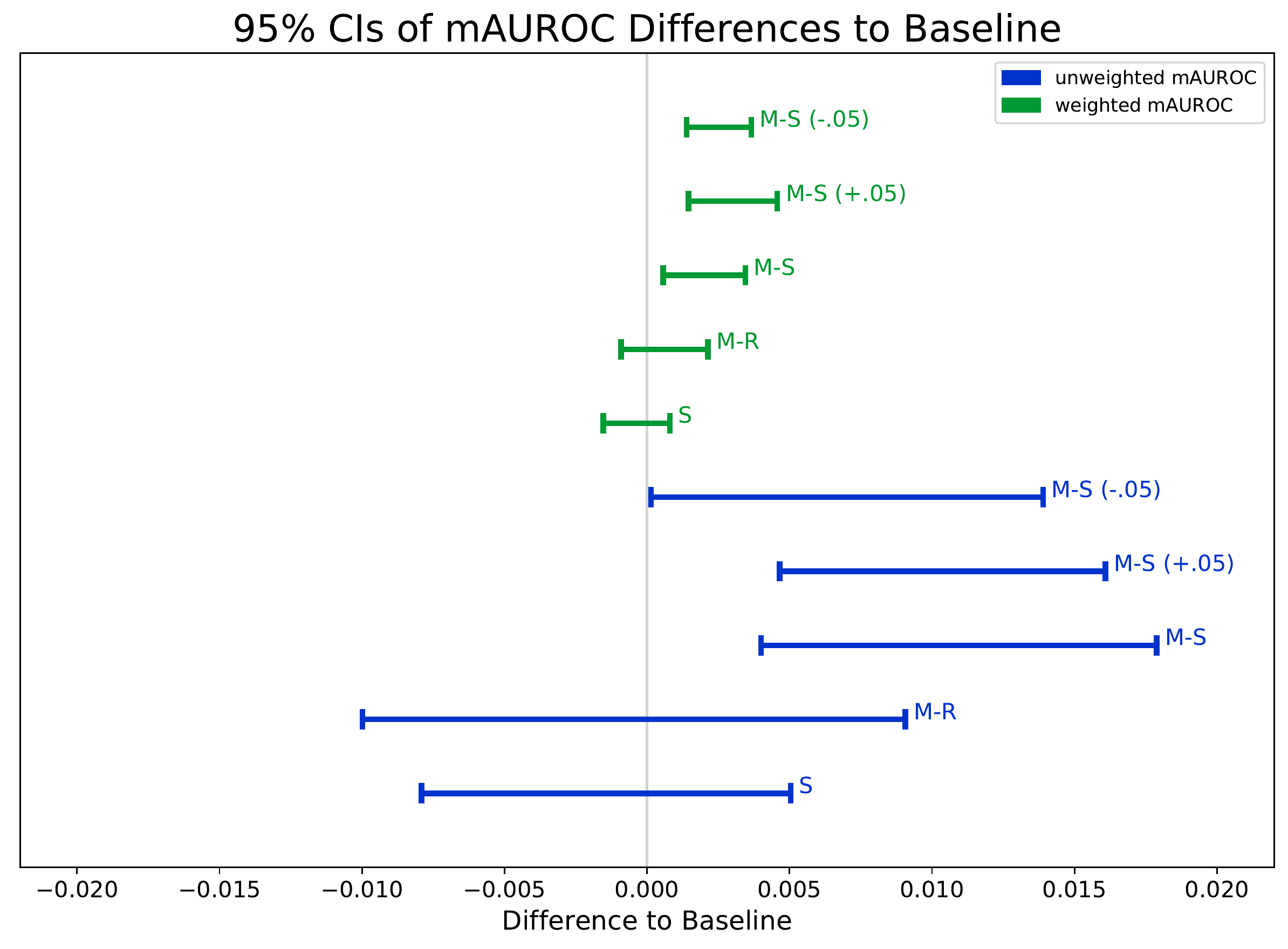}
  \end{center}
  \caption{95\% CIs of the weighted and unweighted mAUROC difference of different experimental setups compared to the baseline \textit{Decay Replay Mining} method.}
  \label{fig:results-predictive}
\end{figure}

\subsection{Predictive Performance}\label{sec:eval:predictive}
This experimental evaluation investigates potential predictive performance improvements of the proposed \textit{Decay Replay Mining} adaptations.
Specifically, five \textit{Decay Replay Mining} adaptations are evaluated against the baseline.
First, a variation of the baseline is investigated which applied the \textit{Swish} activation function modification only. This variation is denoted in the subsequent result overview by \textit{S}. The variation is supposed to provide insight on the impact of the \textit{Swish} activation in the context of \textit{Decay Replay Mining}. 
Second, \textit{M-R} describes the application of the proposed method without the replacing the activation function.
Third, \textit{M-S} denotes the proposed method as described in Section \ref{sec:methodolgy}.
The fourth and fifth model are variations of \textit{M-S} with a 5\% increase and decrease of dropout rate, respectively.
Figure \ref{fig:results-predictive} visualizes the obtained results.

It can be observed that simply replacing the \textit{Rectified Linear Unit} activation function with a \textit{Swish} activation function does not seem to have any impact on the predictive performance of the \textit{Decay Replay Mining} method. Both, the weighted and unweighted mAUROC 95\% CIs do not indicate an improvement.

The results of the proposed approach without the activation function replacement, i.e. \textit{M-R} trend towards marginal improvement of the weighted mAUROC score. Performance increases are not detectable in terms of unweighted mAUROC.

The proposed method denoted by \textit{M-S}, however, indicates predictive performance increases in the ranges of $0.0-0.5$\% and $0.5-1.5$\% in terms of weighted and unweighted mAUROC scores, respectively. While this does not sound like a large increment at first, it needs to be mentioned that the \textit{Decay Replay Mining} approach already delivers an excellent mAUROC score when compared with other state-of-the-art methods.
Hence, an increase of $1.5$\% is significant. 
As expected, the proposed approach supports the balance of minority class predictions as reflected by increasing weighted mAUROC scores. This is the direct result of the masking of the next event using the reachability graph of the PN.

The two dropout rate variations of the proposed methods further increase the performance of the weighted mAUROC. However, the unweighted mAUROC decreases for those models. This can be a sign to further investigate individual hyperparameter optimization per dataset. Nonetheless, the variations further confirm that the proposed approach leads to predictive performance improvements compared to the baseline model.

\subsection{Importance of Precision}\label{sec:eval:precision}
By applying the reachability mask layer to the neural network, one needs to be confident about the quality of the PN.  In the original \textit{Decay Replay Mining} paper, the authors state that the underlying PN should have a high \textit{fitness} score \cite{dreamnap}, i.e. the PN should be able to model all event sequences of a given event log.
In the context of this paper, this means that the true next event is contained in $\mathcal{A}_{M(\tau)}$. If this is not the case, a correct prediction is impossible since the mask layer will cancel out any probability of occurrence of the true next event.
At the same time, the size of $\mathcal{A}_{M(\tau)}$ is desired to be small to provide the neural network with as few options as possible for the next event that still includes the true next event. 
This means that the PN is required to be \textit{fit} and \textit{precise}. 
While this requirement is not trivial and there are ongoing research efforts to obtain highly fitting and precise PNs from event logs \cite{augusto2018automated}, precision can be simulated to experimentally validate the hypothesis of the proposed approach that a higher precision of the PN leads to increased predictive performance.

Two techniques are used to simulate an increased precision of $RG_r$.  
The first one increases the precision by reducing the size of $\mathcal{A}_{M(\tau)}$ in half for each ${M(\tau)}$ given that $|\mathcal{A}_{M(\tau)}| > 2$ using the knowledge of the next event due to the supervised setting of the experiment. 
The resulting reduced set $\mathcal{A}_{M(\tau)}$ is conditioned such that it contains the true next event and has at least a size of 2 unless the original mask had a size of 1. 

The second technique simulates a highly precise $RG_r$ by reducing the size of $\mathcal{A}_{M(\tau)}$ to $2$ given that the initial size is larger than $2$. 
Again, the set reduction is conditioned such that it contains the true next event and has at least a size of 2 unless the original mask had a size of 1. 
In this way, the masks demonstrate high precision while still providing options for the next event. The neural network to predict the next event out of two options provided by the reachability graph rather than out of a large set of all events contained in $\mathcal{A}$.

\begin{figure}[t]
  \begin{center}
    \includegraphics[width=250pt]{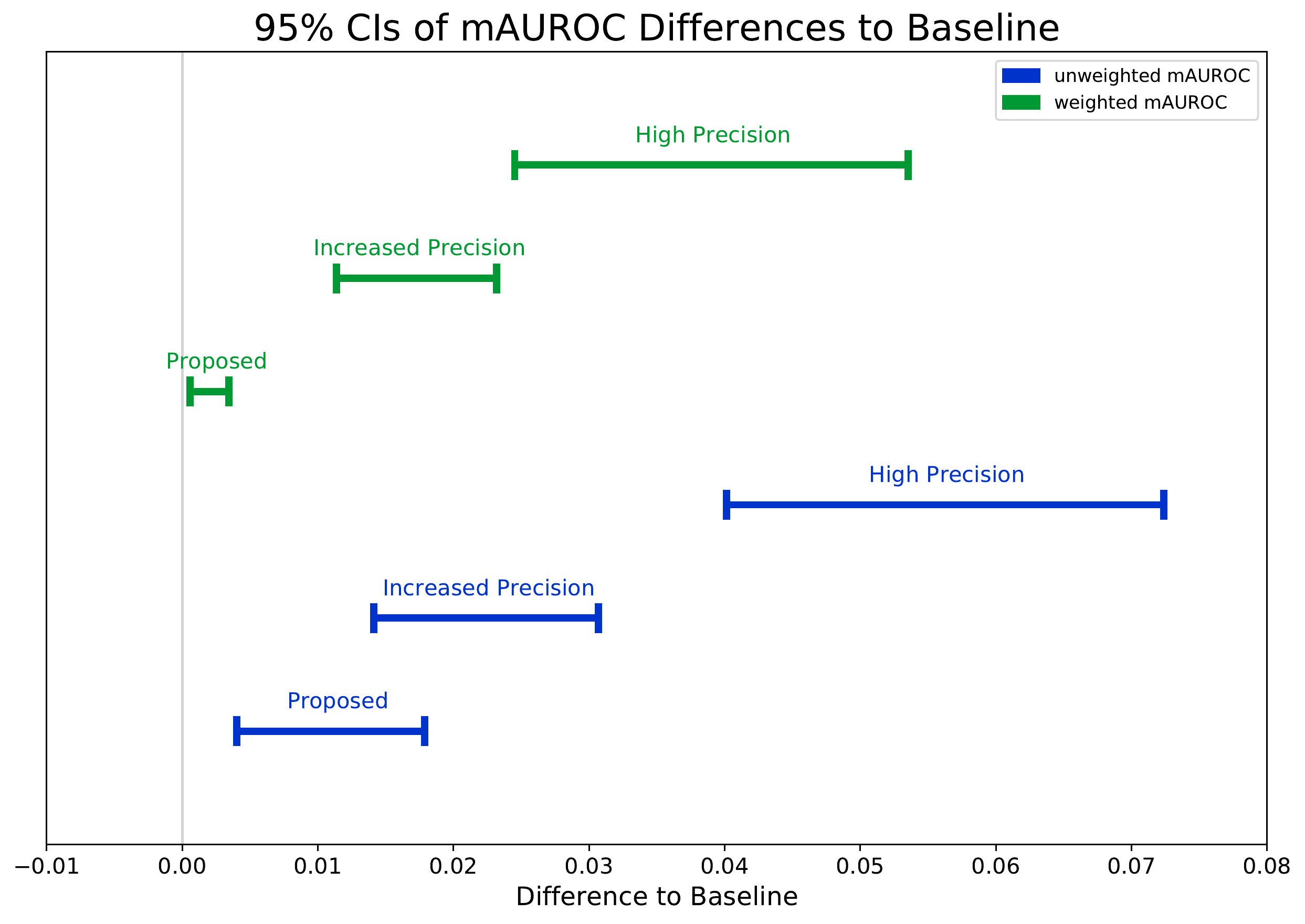}
  \end{center}
  \caption{95\% CIs of the weighted and unweighted mAUROC difference of the proposed method and the proposed method with simulated increased precision and simulated high precision of reachability graphs compared to the baseline \textit{Decay Replay Mining} method.}
  \label{fig:precision-results}
\end{figure}

Both techniques are applied to the proposed approach and are compared to the original baseline. The obtained results are visualized in Figure \ref{fig:precision-results}. As expected, it can be observed that when increasing the precision of the reachability masks, the corresponding weighted and unweighted mAUROC scores increase, too. This further underscores the importance to discover PNs from event logs that reflect the behavior of the system under investigation with a high fitness, high precision, and ideally high generalization.

\section{Conclusion}\label{sec:conclusion}
This paper proposes a method to address the issue of neglecting knowledge that is obtained through process mining efforts for predictive process analytics. 
Specifically, it has been shown how the \textit{Decay Replay Mining} -- a state-of-the-art next event prediction method -- can further interlock a neural network with information from a PN-based process model.
The reachability graph of a PN can be used to mask the outputs of the neural network under the assumption that the process model from which the reachability graph is obtained, is fit and precise. 
An experimental evaluation demonstrated predictive performance gains and highlighted the importance of the process model to be precise.  This work shows that deep learning approaches can be interlocked with explicit process model notations to combine the superior predictive performance advantages with the explainability of process models compared to isolated approaches. 

A limitation of the proposed method is the dependence on the precision and inherently on the generalization of the process model. 
If the PN-based process model allows for a large amount of behavior that does not reflect realistic behavior of the system under investigation, then the corresponding reachability graph is misleading. 
In such a case, applying reachability masking on the neural network will be deceiving. 
Hence, the proposed approach is suggested to be applied carefully if the underlying process model can be trusted. 

Future research is anticipated to conduct a larger set of experiments that investigate the impact of different precision and fitness nuances of underlying PNs on the predictive performance of the proposed method. Additionally, experiments need to be performed on a larger set of benchmark datasets to provide robust statistical conclusions about the effectiveness of the method. 
Moreover, the limitation ineluctably highlights the importance of obtaining trustworthy process models. Therefore, future work is anticipated to continue ongoing research efforts to assess the generalization of process models \cite{polyvyanyy2021bootstrapping, theis2020adversarial} towards the discovery of process models that describe the realistic behavior of the underlying system.



%

\bibliographystyle{IEEEtran}
\bibliography{IEEEabrv,root.bib}

\end{document}